\documentclass[runningheads,a4paper]{llncs}

\usepackage{amssymb}
\usepackage{graphicx}

\usepackage{amssymb}
\usepackage{graphicx}

\usepackage[dvipsnames]{xcolor}
\usepackage{url}

\usepackage{bm}
\usepackage{multirow}
\usepackage{tabularx}
\usepackage{color}
\usepackage{hyperref}
\usepackage{caption}
\usepackage{soul}

\usepackage{booktabs}

\usepackage{times}
\usepackage{epsfig}
\usepackage{graphicx}
\usepackage{amsmath}
\usepackage{amssymb}
\usepackage{comment}
\usepackage{authblk}
\usepackage{multirow}
\usepackage{tabularx}
\usepackage{balance}
\usepackage{subfigure}
\usepackage{balance}
\usepackage{algorithm}
\usepackage{xcolor}
\usepackage{color, colortbl}
\usepackage[noend]{algpseudocode}
\usepackage{cite}
\usepackage{enumitem}
\setlist{nosep}
\usepackage{diagbox}

\usepackage{authblk}

\newcommand{\highlight}[1]{\textcolor{black}{#1}}

\usepackage{pifont}
\newcommand{\cmark}{\ding{51}}%
\newcommand{\xmark}{\ding{55}}%

\def\eg{\emph{e.g}.} 
\def\ie{\emph{i.e}.} 
\def\cf{\emph{c.f}.}

\def\etal{\emph{et~al}.}

\definecolor{lightgray}{gray}{0.9}

\usepackage{url}
\urldef{\mailsa}\path|{ltnghia, nhhuy, jyamagis, iechizen}@nii.ac.jp|    
\newcommand{\keywords}[1]{\par\addvspace\baselineskip
\noindent\keywordname\enspace\ignorespaces#1}

\begin{document}

\mainmatter  

\title{Robust Deepfake On Unrestricted Media: Generation And Detection}

\titlerunning{Robust Deepfake On Unrestricted Media: Generation And Detection}

\author{Trung-Nghia Le\thanks{Corresponding author. This article will appear as one chapter of the book Fake Media Generation and Detection, edited by Mahdi Khosravy, Isao Echizen, Noboru Babaguchi.}
\and Huy H Nguyen \and Junichi Yamagishi \and Isao Echizen}
\authorrunning{Trung-Nghia Le, et al. }
\institute{National Institute of Informatics, Tokyo, Japan\\
The Graduate University for Advanced Studies (SOKENDAI), Kanagawa, Japan\\
University of Tokyo, Tokyo, Japan\\
\mailsa\\
}


\toctitle{Robust Deepfake On Unrestricted Media: Generation And Detection}
\tocauthor{Trung-Nghia Le, et al. }
\maketitle


\begin{abstract}

Recent advances in deep learning have led to substantial improvements in deepfake generation, resulting in fake media with a more realistic appearance. Although deepfake media have potential application in a wide range of areas and are drawing much attention from both the academic and industrial communities, it also leads to serious social and criminal concerns. This chapter explores the evolution of and challenges in deepfake generation and detection. It also discusses possible ways to improve the robustness of deepfake detection for a wide variety of media (\eg, in-the-wild images and videos). Finally, it suggests a focus for future fake media research.

\end{abstract}


\section{Introduction} \label{section:introduction}

Deep learning has been successfully used to power various applications such as big data analysis, natural language processing, signal processing, computer vision, human-computer interaction, medical imaging, and media forensics. Recent advances in deep learning have led to substantial improvements in deepfake generation (``deepfake" for short) (\eg, deep learning-based face forgery, AI-based face forgery) that can change the target person's identity~\cite{faceswap, deepfakes, faceswap-gan, Lingzhi-CVPR2020}. In addition, emerging technologies such as autoencoders (AEs) and generative adversarial networks (GANs) enable transferring one person’s face to another person while retaining the target person’s facial expression and head pose~\cite{Thies-CVPR2016, Thies-TG2019, Nirkin-ICCV2019, Zhixin-ECCV2018}.   

The realistic appearances of images and videos synthesized using deepfake techniques have recently drawn much attention in the computer vision and computer graphics fields. Moreover, they have potential applications in a wide range of areas, such as education~\cite{deepfake_edu}, pattern and design creation, film and art creation (\ie, digital avatars~\cite{Kim-TG2018, Egor-ICCV2019}, and beauty filters~\cite{faceapp, reface, zao}).

However, the ease of creating falsified AI-synthesized images and videos has led to serious concerns about individual harassment, criminal deception, fake news, hoaxes, and financial fraud. Public figures such as celebrities and politicians are easy targets of deepfake attacks. Their faces can be swapped onto the faces of porn stars, for example~\cite{deepfake_porn}. Fake speeches by world leaders can be crafted for falsification purposes, threatening world security. Deepfakes can also be a threat to ordinary people, such as through money scamming~\cite{deepfake_scam}. DeepNude software~\cite{deepnude} is a particularly disturbing threat as it can transform anyone into a non-consensual porn actor. Fake news can cause political and/or religious tensions between countries, fool the public and thereby affect the results of elections, and create chaos in financial markets. These forms of falsification are a massive threat to privacy and identity and can affect many aspects of human lives.

\begin{figure}[t!]
    \centering
    \includegraphics[width=1\linewidth]{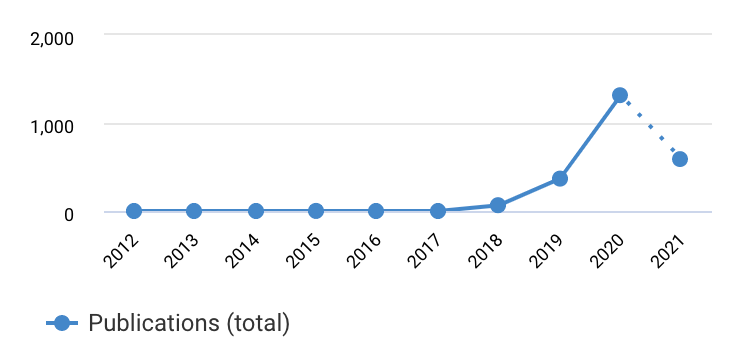}
\caption{Number of deepfake papers published 2012-2021.}
\label{fig:number_papers}
\end{figure}

To address this threat, it is essential to develop countermeasures against face forgeries in digital media. The research community has thus made a great effort to accelerate the development of means for detecting facial manipulation in images and videos. The number of media forensics workshops and conferences has been rapidly increasing. As shown in Figure~\ref{fig:number_papers}, the number of deepfake detection papers\footnote{\url{https://app.dimensions.ai}} has increased substantially in the last few years. In addition, competitions in both academia (NIST’s Open Media Forensics Challenge\footnote{\url{https://mfc.nist.gov/}} and NTU’s DeeperForensics Challenge\footnote{\url{https://competitions.codalab.org/competitions/25228}}) and industry (Facebook’s Deepfake Detection Challenge\footnote{\url{https://www.kaggle.com/c/deepfake-detection-challenge}}) are held regularly to address the threat of face-swapping. 

This chapter provides an in-depth review of deepfake generation and detection methods from the viewpoint of computer vision. It also highlights deepfake datasets and critical benchmarks for different tasks related to face manipulation. In addition, it insightfully discusses potential ways of improving the robustness of deepfake detection methods. 

The remainder of this chapter is organized as follows. Section~\ref{section:generation} introduces deepfake generation methods. Next, Section~\ref{section:detection} presents and discusses deepfake detection methods. Finally, Section~\ref{section:conclusion} summarizes what is covered in this chapter.


\section{Deepfake Generation} \label{section:generation}

\subsection{Deepfake Evolution}


\begin{table}[t!]
\caption{Evolution of deepfake generation methods 2016–2021.}
\label{table:evolution}
\resizebox{1\linewidth}{!}{
\begin{tabular}{|l|l|l|l|l|l|}
\toprule
\multicolumn{1}{|c|}{\textbf{2016}} & \multicolumn{1}{c|}{\textbf{2017}} & \multicolumn{1}{c|}{\textbf{2018}} & \multicolumn{1}{c|}{\textbf{2019}} & \multicolumn{1}{c|}{\textbf{2020}} & \multicolumn{1}{c|}{\textbf{2021}} \\
\midrule
Face2Face~\cite{Thies-CVPR2016}
 & Deepfakes~\cite{deepfakes}
 & FaceSwapGAN~\cite{faceswap-gan}
 & ZAO~\cite{zao}
 & Reface~\cite{reface}
 & Zhu et al.~\cite{Yuhao-CVPR2021}
 \\
 & FaceApp~\cite{faceapp}
 & ProGAN~\cite{Karras-ICLR2018}
 & StyleGAN~\cite{Karras-CVPR2019}
 & DeepFaceLab~\cite{Ivan-2020}
 & Le et al.~\cite{ltnghia-ICCV2021}
 \\
 & Korshunova et al.~\cite{Korshunova-ICCV2017}
 & StarGAN~\cite{Choi-CVPR2018}
 & NeuralTextures~\cite{Thies-TG2019}
 & StyleGANv2~\cite{Karras-CVPR2020}
 &  \\
 &  & ReenactGAN~\cite{Wayne-ECCV2018}
 & GANimation~\cite{Pumarola-IJCV2019}
 & StarGANv2~\cite{Choi-CVPR2020}
 &  \\
 &  & RsGAN~\cite{Natsume-SIGGRAPH2018}
 & FSGAN~\cite{Nirkin-ICCV2019}
 & InterFaceGAN~\cite{Shen-CVPR2020}
 &  \\
 &  & X2Face~\cite{Wiles-ECCV2018}
 & Zhang et al.~\cite{Zhang-BMVC2019}
 & StyleALAE~\cite{Pidhorskyi-CVPR2020}
 &  \\
 &  & FSNet~\cite{Natsume-ACCV2018} & Egor et al. \cite{Egor-ICCV2019}
 & FaceShifter~\cite{Lingzhi-CVPR2020}
 &  \\
 &  & Zhixin et al.~\cite{Zhixin-ECCV2018}  &  & ICface \cite{Soumya-WACV2020}
 &  \\
 &  & Kim et al.~\cite{Kim-TOG2018}
 &  & FaR-GAN~\cite{Hanxiang-CVPRW2020}
 & \\
 \bottomrule
\end{tabular}
}
\end{table}

Table \ref{table:evolution} illustrates the evolution of deepfakes from 2016 to 2021. Traditional manipulation methods based on hand-crafted features were developed to add (splice), remove (inpaint), and replicate (copy/move) objects within images. Recent advancements in deep learning support various automated manipulation approaches that achieve realistic appearances, namely deepfake generation. The first deepfake generation method developed was based on a simple AE~\cite{deepfakes}. Subsequently, GANs~\cite{faceswap-gan} started to be used to create realistic fake media. GAN-based methods are more generalized than AE-based ones in synthesizing realistic manipulated faces because they work without being explicitly trained on subject-specific images. Now, AE- and GAN-based methods are the two most widely used deepfake generation methods.

Deepfake generators are usually categorized by their application (\eg, identity swapping, expression reenactment, face synthesis, and attribute manipulation). Identity swapping replaces the target identity with the source identity; expression reenactment manipulates facial expressions; face synthesis creates fake facial images; attribute manipulation creates attribute fabrication. In addition, here we also differentiate methods by the evolution of the techniques they use. Furthermore, we analyze their pros and cons. We argue that modern deep learning techniques can be used in a variety of areas.

Paired-training techniques~\cite{deepfakes}, first proposed in 2017, have the highest ratio of usage in deepfake generators. They focus on specific individuals and require a substantial amount of data on both the target and source individual. Moreover, the use of paired training for deepfake generation is a time-consuming process. The evolution of deep learning techniques has recently enabled the development of few-/one-/zero-shot learning approaches that overcome these disadvantages. 

\subsection{Identity Swapping}

In identity swapping, or face replacement, a person's face in a source image is replaced with another person's face. This manipulation approach uses a deep generator (\ie, AE or GAN) to give a victim’s face someone else’s features while preserving the original facial expression.

The first deep learning method for identity swapping, Deepfakes~\cite{deepfakes}, uses AEs and was developed in 2017. It has been used, for example, to replace the faces of actors in pornographic videos with those of celebrities. Two AEs (\ie, encoder-decoder pairs) are used in which the encoder is used to extract the latent features of a face and the decoder is used to reconstruct the face. After each encoder-decoder pair is trained on a person’s video, the decoders are swapped to regenerate the target face with the features of the source face.

Korshunova \etal~\cite{Korshunova-ICCV2017} used a fully convolutional network to transfer the facial appearance of the source person to that of another person in another image. Utilizing the style transfer technique, they combined style loss, content loss, light loss, and total variation regularization to produce realistic images. Open-source projects (\eg, DeepFaceLab~\cite{Ivan-2020} and FaceSwap-GAN~\cite{faceswap-gan}) with tutorials available on GitHub have opened the door for anyone to create fake images and videos. However, these approaches require a large amount of data on both the target and source individual for paired training, resulting in a time-consuming process. 

The time-consuming limitation has been partially overcome by recent developments that have made the generation of deepfakes more efficient. Zakharov \etal~\cite{Egor-ICCV2019} proposed using GAN-based few-/one-shot learning to generate a realistic talking-head video from an image, but this requires lengthy meta-learning on a large-scale video dataset. Li \etal~\cite{Lingzhi-CVPR2020} trained a two-stage framework in a zero-shot learning manner (\ie, self-supervised learning) for high fidelity and occluded face-swapping. Zhu \etal~\cite{Yuhao-CVPR2021} extended the latent space to maintain more facial details and then used StyleGAN2~\cite{Karras-CVPR2020} to generate high-resolution swapped facial images. Natsume \etal~\cite{Natsume-ACCV2018} combined separately encoded face and facial landmarks to generate a fake identity. These methods can be used to swap any two faces without retraining. Commercial deepfake applications launched since 2017 (\eg, FaceApp~\cite{faceapp}, Reface~\cite{reface}, ZAO~\cite{zao}) enable users without technical knowledge to easily swap faces in images and videos. However, they cannot handle faces with pose variations or that are partly occluded.

\subsection{Expression Reenactment}

Expression reenactment, also known as face reenactment or emotion synchronization, manipulates facial expressions by transferring facial expressions, gestures, and head movements of the source person to the target person while keeping the identity of the target person. This kind of manipulation is usually aimed at altering the person's facial expressions and synchronizing lip movements to create fabricated content.

Facial expression reenactment techniques are mainly based on 3D face reconstruction and a GAN architecture. The 3D facial modeling-based approach can accurately capture the geometry and movement of the head, resulting in a photorealistic reenacted face. Thies \etal~\cite{Thies-CVPR2016, Thies-TG2019} introduced the use of 3D facial modeling combined with image rendering for real-time transfer of the facial expressions of a person talking in front of a commodity webcam into a target face video. 

Although GAN-based methods can generate photorealistic images, they require a large amount of training data in order to achieve photorealistic reenactment for unknown identities. Wu \etal~\cite{Wayne-ECCV2018} proposed encoding the source face into a boundary latent space and using a target-specific decoder to transfer the latent features onto the target face. Shu \etal~\cite{Zhixin-ECCV2018} combined the disentangling of shapes in an unsupervised manner with AE models to morph expressions. Kim \etal~\cite{Kim-TOG2018} proposed using space-time encoding for temporally coherent synthesis in combination with a conditional GAN (cGAN) to synthesize video portraits of a target individual from a still image. This enables full control over the target by transferring a rigid head pose, facial expression, and eye motion with a high level of photorealism. Pumarola \etal~\cite{Pumarola-IJCV2019} trained emotion action units based on a dual-cGAN in a fully unsupervised manner for use in generating facial animation from a single image. In addition to transferring expressions, the head pose can be controlled by using a recurrent neural network to enhance naturalness~\cite{Nirkin-ICCV2019} by using different modalities~\cite{Wiles-ECCV2018} and by using human interpretable attributes and actions~\cite{Soumya-WACV2020}.

Few-/one-shot facial expression reenactment methods have recently been introduced to overcome the disadvantage of training on large-scale data by using a few- or even a single-source image. Ha \etal~\cite{Ha-AAAI2020} proposed using a few-shot face reenactment framework that includes image attention analysis, target feature alignment, and landmark transformation to prevent quality degradation in unseen identity mismatch situations. These methods do not need additional fine-tuning phases for identity adaptation, yielding deepfake images that can be usefully deployed in the wild. SPADE blocks~\cite{Park-CVPR2019}, used in one-shot face reenactment, can effectively generate a new source’s expressed appearance by learning the latent representation of the target's facial landmarks~\cite{Zhang-BMVC2019, Hanxiang-CVPRW2020}. The problem has also been extended to multimedia-based reenactment. Fried \etal~\cite{Fried-TG2019} developed a text-based editing approach to generating talking-head videos—the dialogue is modified to match the corresponding head movement of the speaker.

\subsection{Face Synthesis}

Face synthesis has been applied in many fields, such as video games and 3D modeling. Almost all face synthesis methods utilize powerful GAN models to generate entirely new facial images. GAN models have been developed that have progressively improved resolution, image quality, and realism. StyleGAN~\cite{Karras-CVPR2019} improved the image resolution ($1024 \times 1024$) of ProGAN~\cite{Karras-ICLR2018} ($128 \times 128$) while StyleGAN2~\cite{Karras-CVPR2020} further improved the image quality of StyleGAN by removing unwanted artifacts. 

GAN-based face synthesis methods have been used in various applications, such as for facial attribute translation~\cite{Choi-CVPR2018, Choi-CVPR2020, Karras-CVPR2019, Karras-CVPR2020}, identity-attribute combination~\cite{Bao-CVPR2018}, identified characteristics removal~\cite{Maximov-CVPR2020}, and interactive semantic manipulation~\cite{ChengHan-CVPR2020, Zhu-CVPR2020}. In addition, some GAN-based methods can be used in virtual try-on applications that enable consumers to try on various cosmetics in a virtual environment~\cite{Wentao-CVPR2020, Nguyen-CVPR2021}. These methods can be extended to the entire body. For example, DeepNude uses the Pix2PixHD GAN model~\cite{Wang-CVPR2018} to inpaint clothing regions to generate fake nude images.

\subsection{Attribute Manipulation}

Facial attribute manipulation, also known as face editing or face retouching, modifies facial attributes such as hair color, hairstyle, skin color, gender, age, smile, eyeglasses, and makeup. This type of manipulation can be considered partially conditional face synthesis. Thus, GAN techniques for face synthesis are usually used for facial attribute manipulation. Choi \etal~\cite{Choi-CVPR2018, Choi-CVPR2020} proposed using a unified model architecture that enables simultaneous training of multiple datasets with different domains to transfer various facial attributes and expressions at the same time. Disentangled facial features can be interpreted in different latent spaces, resulting in more precise control of attribute manipulation in face editing~\cite{Karras-CVPR2019, Karras-CVPR2020, Shen-CVPR2020, Pidhorskyi-CVPR2020}. These GAN-based methods are agnostic and trained on a huge number of images in order to work without being explicitly trained on subject-specific images. However, they tend to generate faces with new identities if the target face is not in the training distribution. In addition, they cannot handle faces with pose variations or occlusions.

\subsection{Hybrid Applications}

We argue that current deep learning methods can combine different deepfakes to create hybrid applications. Nirkin \etal~\cite{Nirkin-ICCV2019} introduced a GAN model for both face-swapping and reenactment in real time that follows the reenact and blend strategy. Natsume \etal~\cite{Natsume-SIGGRAPH2018} encoded latent vectors for face and hair as two separate variational AEs (VAEs) and then conditionally swapped or edited the identity of the target. Le \etal~\cite{ltnghia-ICCV2021} proposed a framework for generating fake identities by combining facial attribute editing and face-swapping. These methods are subject agnostic and can be applied to any pair of faces without retraining. Commercial FaceApp mobile applications~\cite{faceapp} enable users to swap faces as well as change the age, gender, smile, and hairstyle of the faces.


\subsection{Limitations}

\begin{figure}[t!]
    \centering
    \includegraphics[width=1\linewidth]{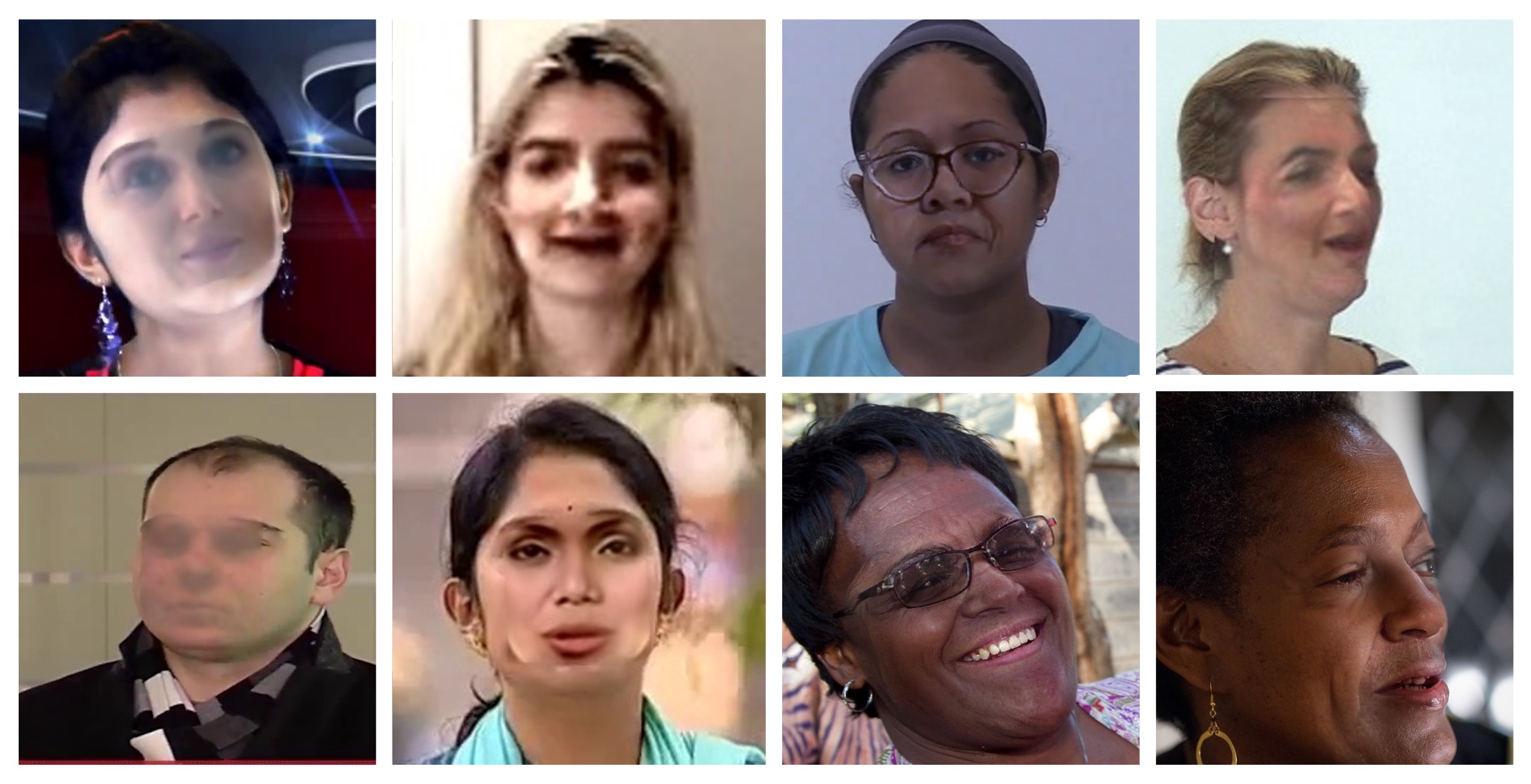}
\caption{Problems with current deepfake generation methods. From left to right: low resolution, low quality, strange artifacts due to wearable items, and facial pose variations.}
\label{fig:limitations}
\end{figure}

The biggest limitation of deepfake methods is the \textbf{lack of generalizability}, which can lead to such problems as low resolution, low quality, strange artifacts, facial pose variations, and occlusions (\cf~Fig.~\ref{fig:limitations}).

Images generated using conventional deepfake methods usually have \textbf{low resolution and low quality}. While some recently developed methods have improved resolution and image quality, they require a large amount of training data (\ie, more than 1.5 million images), leading to time-consumed training. 

\textbf{Strange artifacts} in generated images are mainly caused by ineffective blending and facial pose variations. Although deepfake methods for generating fake faces have steadily improved, the mismatch between the synthesized regions and original regions mostly appears due to the limitations of the blending algorithms. Hence, analysis of undesired colors and patterns in blending boundaries around the face can lead to strong deepfake detection~\cite{Li-CVPR2020}, even for unknown deepfake generators.

\textbf{Facial pose variations}, which generally affect in-the-wild images, usually cannot be handled with non-paired-training methods. Existing deepfake methods are generally trained on frontal view facial images; they thus fail to preserve the face when there are large pose variations. Although the improved blending with recently developed methods~\cite{ltnghia-ICCV2021, Lingzhi-CVPR2020} helps to reduce artifacts, it does not completely solve the problem. 

\textbf{Occlusions} caused by hair, glasses, or other objects cannot be handled by conventional deepfake generation methods. This leads to visual artifacts in the synthesized fake faces. A recently developed heuristic error acknowledging refinement network~\cite{Lingzhi-CVPR2020} helps to remove facial inconsistencies.



\section{Deepfake Detection} \label{section:detection}

Although recently developed deepfake generation methods generate realistic images that can spoof machines and fool people, visual anomalies in the images, such as low quality, low resolution, and strange artifacts, are still present. This means that differences between the genuine and manipulated areas can be parameterized and used for deepfake detection.

\begin{figure}[t!]
    \centering
    \begin{tabularx}{\linewidth}{*{3}{X}}
        \hfill\includegraphics[width=\linewidth]{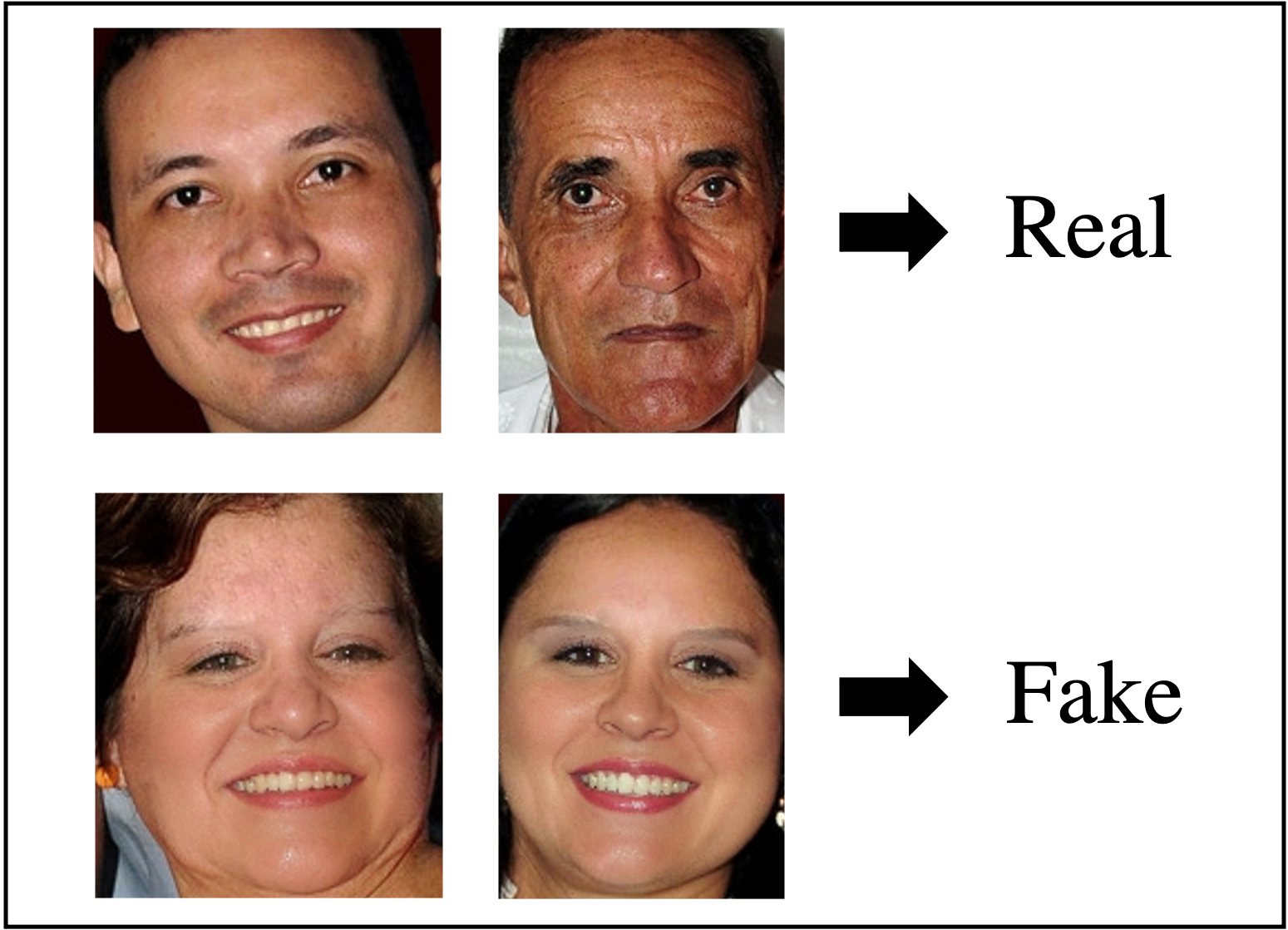}\hspace*{\fill} &
        \hfill\includegraphics[width=\linewidth]{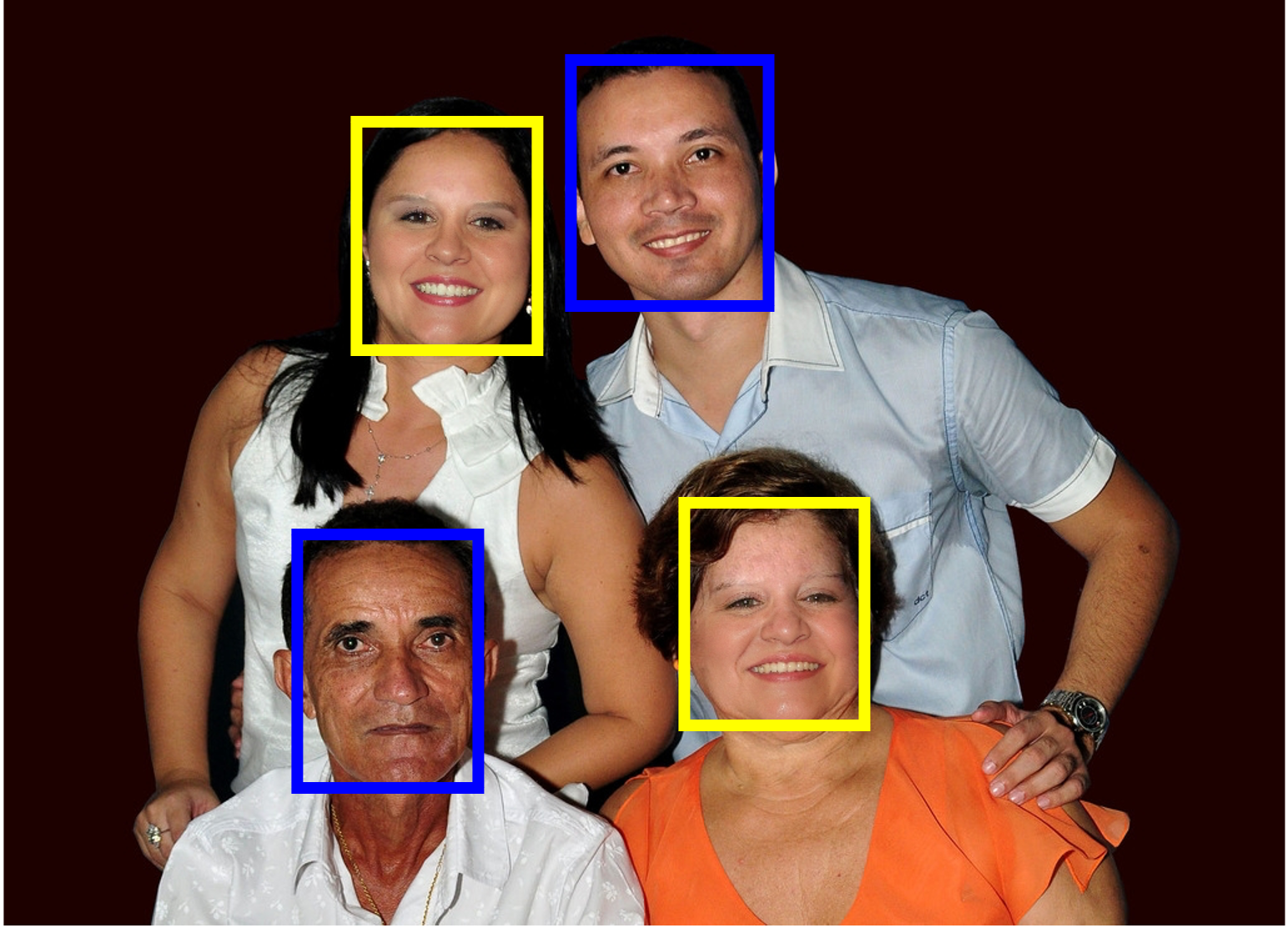}\hspace*{\fill} &
        \hfill\includegraphics[width=\linewidth]{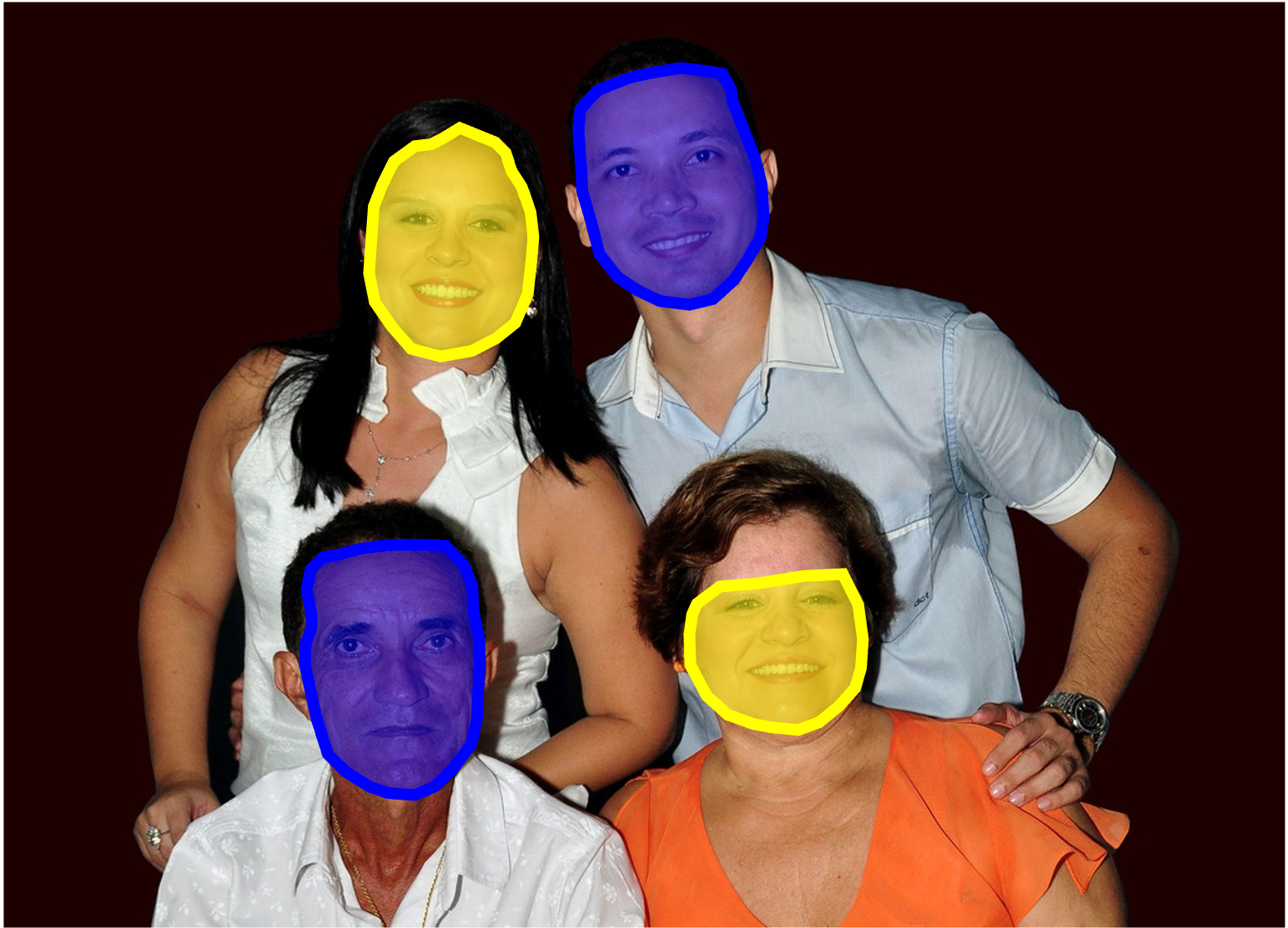}\hspace*{\fill} \\
        \centering \scriptsize{a) Classification} &
        \centering \scriptsize{b) Detection} &
        \centering \scriptsize{b) Segmentation} 
    \end{tabularx}
    \caption{Deepfake tasks (from left to right): conventional classification, end-to-end detection, and segmentation.}
    \label{fig:deepfake_category}
\end{figure}

Conventional deepfake detection methods can only classify a given cropped face as real or fake\footnote{In this chapter, we use "deepfake detection" following existing work, but the actual task is “recognition/classification.}. Recently developed methods can not only recognize the authenticity of faces but also localize manipulated areas at the bounding-box and pixel levels, namely detection and segmentation, respectively (\cf~Fig.~\ref{fig:deepfake_category}).
 
\begin{figure}[t!]
    \centering
    \begin{tabularx}{\linewidth}{*{1}{X}}
        \hfill\includegraphics[width=0.9\linewidth]{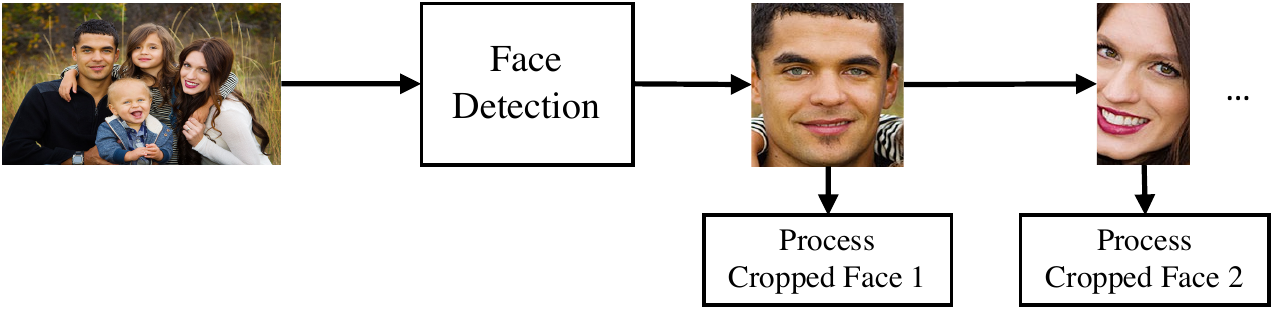}\hspace*{\fill} \\
         \centering a) Conventional approach \\  
         \hfill \\
        \hfill\includegraphics[width=0.9\linewidth]{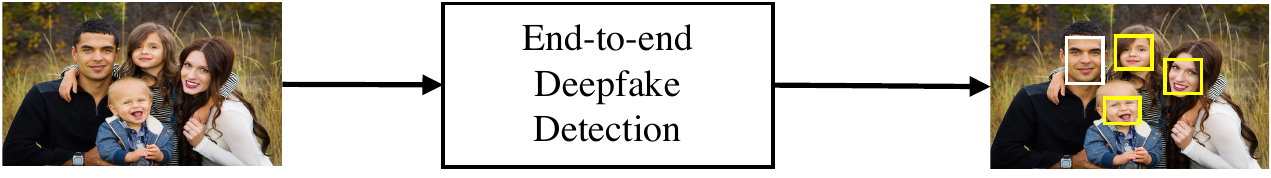}\hspace*{\fill} \\
        \centering b) End-to-end approach
    \end{tabularx}
    \caption{Two basic approaches to deepfake detection: methods based on conventional approach use sequential processing while those based on end-to-end approach use parallel processing.}
    \label{fig:deepfake_detection}
\end{figure}

This section describes the two basic approaches to deepfake detection, the conventional approach and the end-to-end approach (\cf~Fig.~\ref{fig:deepfake_detection} and Table~\ref{table:deepfake_detection}). It also analyzes the robustness of deepfake detection methods and suggests future research directions. 


\begin{table}[t!]
\caption{Categorization of deepfake detection methods. $\bigstar, \clubsuit,$ and $\spadesuit$ indicate classification, detection, and segmentation tasks, respectively. Methods that can process multiple faces are shown in \textcolor{blue}{blue}.}
\label{table:deepfake_detection}
\resizebox{1\linewidth}{!}{
\begin{tabular}{|l|l|l|l|l|}
\toprule
\multicolumn{4}{|c}{\textbf{Conventional Approach}} & \multicolumn{1}{|c|}{\textbf{End-to-end Approach}} \\
\cmidrule{1-4}
\multicolumn{1}{|c|}{\textbf{Data-Driven}} & \multicolumn{1}{c|}{\textbf{Visual Artifacts}} & \multicolumn{1}{c|}{\textbf{Frequency Domain}} & \multicolumn{1}{c}{\textbf{Attention \& Segmentation}} & \multicolumn{1}{|c|}{\textbf{}} \\
\midrule
Zhou~\etal~\cite{Zhou-CVPRW2017} $\bigstar$ & Li~\etal~\cite{Li-WIFS2018} $\bigstar$ & Li~\etal~\cite{Li-CVPR2021} $\bigstar$ & Nguyen~\etal~\cite{nhhuy-BTAS2019} $\bigstar, \spadesuit$ & \textcolor{blue}{Zhou~\etal~\cite{Zhou-CVPR2021}} $\bigstar, \clubsuit$ \\
Afchar~\etal~\cite{Afchar-WIFS2018} $\bigstar$ & Matern~\etal~\cite{Matern-WACVW2019} $\bigstar$ & Liu~\etal~\cite{Liu-CVPR2021} $\bigstar$ & Dang~\etal~\cite{Dang-CVPR2020} $\bigstar$ & \textcolor{blue}{Le~\etal~\cite{ltnghia-ICCV2021}} $\bigstar, \clubsuit, \spadesuit$ \\
Nguyen~\etal~\cite{nhhuy-ICASSP2019} $\bigstar$ & Yuezun~\etal~\cite{Yuezun-CVPRW2019} $\bigstar$ &  & Li~\etal~\cite{Li-CVPR2020} $\bigstar, \spadesuit$ &  \\
Rossler~\etal~\cite{Rossler-ICCV2019} $\bigstar$ & Yang~\etal~\cite{Yang-ICASSP2019} $\bigstar$ &  & Bojia~\etal~\cite{Bojia-MM2020} $\bigstar$ &  \\
Wang~\etal~\cite{Wang-IJCAI2020} $\bigstar$ & Haliassos~\etal~\cite{Haliassos-CVPR2021} $\bigstar$ &  & Wang~\etal~\cite{Wang-CVPR2021} $\bigstar$ &  \\
 & Zhu~\etal~\cite{Zhu-CVPR2021} $\bigstar$ &  &  &  \\
\bottomrule
\end{tabular}
}
\end{table}


\subsection{Conventional Deepfake Detection}

Face forgery classification, \ie, conventional deepfake detection, is aimed at classifying facial images as real or fake. Face forgery classification methods require the input of identified face regions (\cf~Fig.~\ref{fig:deepfake_detection}a). They do not have a face localization ability, so their performance greatly depends on the accuracy of the independent face detection method used. 

Conventional deepfake classifiers are data-driven; their deep networks are directly trained on real and fake images and videos without relying on specific artifact analysis, such as a two-stream neural network~\cite{Zhou-CVPRW2017}, MesoNet~\cite{Afchar-WIFS2018}, CapsuleNet~\cite{nhhuy-ICASSP2019}, XceptionNet~\cite{Rossler-ICCV2019}, and FakeSpotter~\cite{Wang-IJCAI2020}.

Some methods exploit inconsistencies created by visual artifacts in deepfake images and videos; they analyze biological clues such as eye blinking~\cite{Li-WIFS2018}, head pose~\cite{Yang-ICASSP2019}, skin texture~\cite{Liu-CVPR2020}, and iris and teeth color~\cite{Matern-WACVW2019}. Other methods use artifacts in affine face warping~\cite{Yuezun-CVPRW2019} or in mouth movements~\cite{Haliassos-CVPR2021} to distinguish real and fake faces. Another method~\cite{Zhu-CVPR2021} decomposes the facial image to extract facial details (\ie, direct light and identity texture) and combines them with the original face through a two-stream network to find critical forgery clues.

The frequency domain has been used in a few methods, which improves their transferability. Li \etal~\cite{Li-CVPR2021} developed an adaptive frequency feature generation module to mine frequency clues in combination with metric learning for improved separability in the embedding space. Liu \etal~\cite{Liu-CVPR2021} combined the spatial image and phase spectrum to capture up-sampling artifacts and thereby improve the transferability of deepfake detection. 

Most current deepfake classifiers are based on attention or segmentation masking. Wang \etal~\cite{Wang-CVPR2021} presented an attention-based data augmentation method for guiding the detector to refine and enlarge its attention mask by occluding sensitive facial regions. Zi \etal~\cite{Bojia-MM2020} fed both the facial image and attention mask (\ie, blurred facial landmark regions) into an attention-based face fusion network for improved classification. Dang \etal~\cite{Dang-CVPR2020} used an attention mechanism to highlight informative regions and thereby further improve feature maps used for the classification task. Li \etal~\cite{Li-CVPR2020} focused on blending regions to enable forged faces to be recognized without relying on knowledge of specific face manipulation techniques. Nguyen \etal~\cite{nhhuy-BTAS2019} designed a multi-task learning network for simultaneously locating manipulated regions in fake facial images.

\highlight{Most existing methods aim to identify the authenticity of faces without categorizing deepfake generation types. The main reasons are the lack of data and ambiguity between deepfake categories. Recently introduced methods, which are trained on datasets~\cite{Rossler-ICCV2019} with multiple deepfake types, generally have the ability to recognize deepfake generation types~\cite{nhhuy-ICASSP2019, Rossler-ICCV2019, Wang-IJCAI2020, Dang-CVPR2020, Haliassos-CVPR2021, Zhu-CVPR2021, Li-CVPR2021, Liu-CVPR2021}. However, these methods lack generalization due to overfitting to few generation methods in the training process and thus cannot be used in practical contexts.}


\subsection{End-to-end Deepfake Detection}

Detecting forged faces among multiple faces in multi-person scenes, in which only a small subset of them have been manipulated, is still challenging for traditional deepfake detection methods. Several researchers have recently begun to target multi-person in-the-wild images (\cf~Fig.~\ref{fig:deepfake_detection}b). Zhou \etal~\cite{Zhou-CVPR2021} trained an attention framework to detect face forgeries in multi-person scenes. Le \etal~\cite{ltnghia-ICCV2021} used transfer learning to build end-to-end models for multiple face forgery detection and segmentation for in-the-wild images. 


\subsection{Robust Deepfake Detection} \label{section:robustness}

\subsubsection{Vulnerability to Adversarial Attacks}

Adversarial attacks can modify images and video by adding specific noises in such a way that machine learning models misclassify them. CNNs are vulnerable to such attacks, and their accuracy may drop to near zero~\cite{Carlini-CVPRW2020}. Indeed, the addition of insignificant noise can cause remarkable changes in the prediction and even completely confuse deep learning classifiers. Hussain \etal~\cite{Hussain-WACV2021} adversarially modified fake videos and presented pipelines in both white-box and black-box attack scenarios that can fool deepfake classifiers into classifying fake videos as real. Huang \etal~\cite{Huang-ICIP2020} showed the existence of both individual and universal adversarial perturbations that can cause well-performing deepfake classifiers to misbehave. Li \etal~\cite{Dongze-CVPR2021} demonstrated that fake facial images generated using adversarial points on a face manifold can defeat two strong forensic classifiers. Even methods that won the Deepfake Detection Challenge (DFDC)~\cite{Dolhansky-2020} were easily bypassed in a practical attack scenario using transferable and accessible adversarial attacks~\cite{Neekhara-CVPRW2021}. 

Object detectors are more robust than classifiers due to their complex network architecture, but there have been few reports on attacking such detectors. Treu \etal~\cite{Marc-CVPRW2021} developed a robust adversarial example method that degrades human detectors by targeting clothing regions.

Defense against adversarial attacks is a critical step toward developing robust solutions for biometrics verification (\ie, deepfake detection). Many solutions have been proposed for increasing the robustness of CNNs against adversarial attacks. Most of them are independent solutions and thus do not require retraining on the victim images. Naseer \etal~\cite{Naseer-CVPR2020} introduced a self-supervised countermeasure against unseen adversarial attacks that can be deployed as a plug-and-play solution to protect different vision systems, such as those for classification, segmentation, and detection. Salman \etal~\cite{Salman-NeurIPS2020} presented a defense method called ``denoised smoothing" that converts existing pre-trained classifiers into provably robust ones without any retraining or fine-tuning. This method can be easily integrated into public vision API providers (\ie, Azure, Google, AWS, and Clarifai APIs). To defend black-box face biometrics classifiers against adversarial attacks, Theagarajan \etal~\cite{Theagarajan-CVPRW2020} used an ensemble of iterative adversarial image purifiers for which the performance is continuously validated in a loop using Bayesian uncertainties. Akhil \etal~\cite{Akhil-CVPRW2020} presented a parameter-free defense layer that is plugged into a CNN to prevent gradient- and optimization-based adversarial attacks in black-box and gray-box settings. In addition, AE- and GAN-based denoising methods~\cite{Moran-CVPR2020} have been used to reconstruct the original images.

\subsubsection{Nonrobustness on Unseen Scenarios}

Most conventional deepfake detection methods greatly depend on the training scenario. They are generally trained on a few of the more widely used deepfake generation methods, which are applied to a limited number of images and videos. This training of deepfake detection models on similar data can lead to overfitting poor performance in the real world. Furthermore, conventional deepfake detection methods are not robust against media processing, such as compression, noise addition, color variation, and light variation. Indeed, although strong deep learning models have attained very high accuracy~\cite{nhhuy-ICASSP2019, Li-CVPR2020} in laboratory environments with a simple background and a single clear front-view face~\cite{Korshunov-2018, Yang-ICASSP2019}, they generally do not work well in diverse contexts that do not share a close distribution with the training dataset. Le \etal~\cite{ltnghia-ICCV2021} showed that even end-to-end deepfake detectors are easily degraded on unseen image scenarios, \ie, ones distant from the training distribution.

To deal with unseen forgery types, Haliassos \etal~\cite{Haliassos-CVPR2021} targeted inconsistencies in mouth movements by leveraging rich representations learned via lipreading. Li \etal~\cite{Li-CVPR2020} focused on blending regions to recognize forged faces without relying on knowledge of specific face manipulation techniques. Sun \etal~\cite{Sun-CVPR2021} developed a temporal method for geometrically modeling discriminative features to improve robustness for highly compressed or noise corrupted videos. 

Data augmentation has also been shown to be effective for training robust deep networks. Chen \etal~\cite{Chen-CVPR2021} augmented the process of training general object detectors. They explored model-dependent data augmentation by dynamically selecting more substantial and relevant adversarial images sourced from classification and localization branches, which can be generalized to different object detectors. In addition, improvements have been made to a couple of existing datasets by using simple perturbations, which have increased their size, such as the DFDC~\cite{Dolhansky-2020} and DeeperForensics~\cite{Jiang-CVPR2020} datasets. Le \etal~\cite{ltnghia-ICCV2021} imitated real-world data distributions through diverse perturbations, such as color and edge manipulation, block-wise distortion, image corruption, and weather effects. Hence, to deal with unseen scenarios, recently developed augmentation libraries can be used to better train robust deepfake detection methods. In addition, if the new contextual images are still distant from the training distribution, a budget-wise solution is to use CycleGAN~\cite{Zhu-ICCV2017} to simulate unseen real-world contexts for training robust deep learning models.

\begin{figure}[t!]
\centering
\includegraphics[width=1\linewidth]{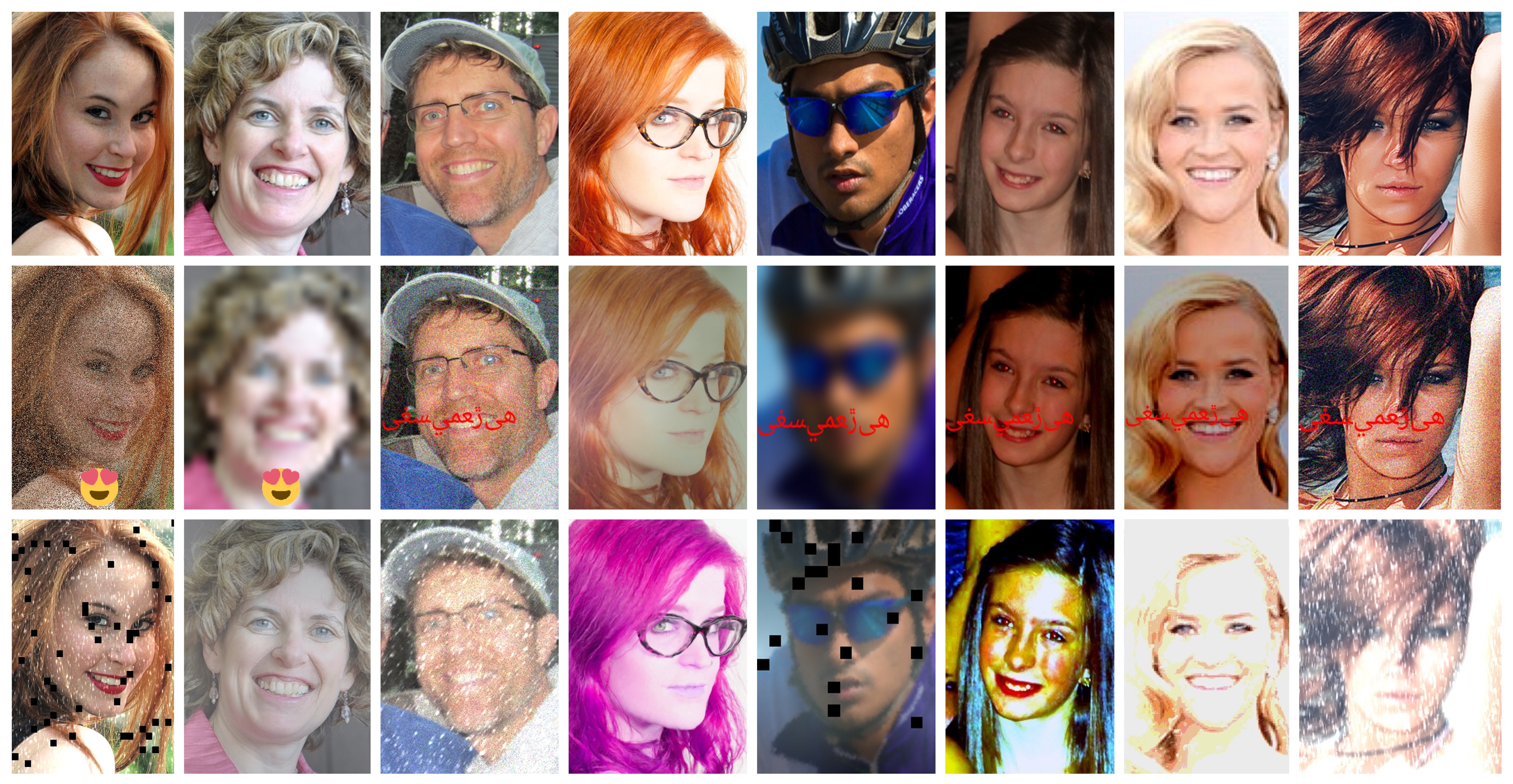}
\caption{Images synthesized using robust deepfake detection methods that were trained using augmentation libraries. From top to bottom, original images, images generated using AugLy~\cite{AugLy-2021}. and images generated using OpenForensics~\cite{ltnghia-ICCV2021}.}
\label{fig:augmentation}
\end{figure}


\subsection{Deepfake Datasets and Benchmarks}

This section introduces deepfake datasets and analyzes methods that attained a top ranking in recent deepfake challenges.

\subsubsection{Deepfake Datasets}

\begin{table}[t!]
\caption{Basic information about deepfake datasets. “Cls.,” “Det.,” and “Seg.” stand for classification, detection, and segmentation, respectively. Pristine scenarios are originally collected images/videos used to generate fake data. Unique fake scenarios are fake images/videos ignoring perturbations.}
\label{table:deepfake_datasets}
\resizebox{1\linewidth}{!}{
\begin{tabular}{|l|c|c|c|c|c|c|c|c|c|}
\toprule
\textbf{Dataset} & \textbf{Year} & \textbf{Type} & \textbf{Task} & \textbf{Fake Identity} & \textbf{Multi-Face} & \begin{tabular}[c]{@{}c@{}}\textbf{Face}\\\textbf{Occlusion}\end{tabular} & \begin{tabular}[c]{@{}c@{}}\textbf{No. of Pristine}\\\textbf{Scenario}\end{tabular} &
\begin{tabular}[c]{@{}c@{}}\textbf{No. of Unique Fake}\\\textbf{Scenario}\end{tabular} & \begin{tabular}[c]{@{}c@{}}\textbf{Data}\\\textbf{Augmentation}\end{tabular} \\
\midrule
DF-TIMIT~\cite{Korshunov-2018} & 2018 &  Video  & Cls. & Other videos & \xmark & \xmark & 320 & 320 & \xmark \\
UADFV~\cite{Yang-ICASSP2019} & 2019 &  Video  & Cls. & Other videos & \xmark & \xmark & 49 & 49 & \xmark \\
FaceForensics++~\cite{Rossler-ICCV2019} & 2019 & Video  & Cls. & Other videos & \xmark & \xmark & 1,000 & 4,000 &  \xmark \\
Google DFD~\cite{google_dfd-2019} & 2019 &  Video  & Cls. & Other videos & \xmark & \xmark & 363 & 3,068  & \xmark \\
Facebook DFDC~\cite{Dolhansky-2020}  & 2020 & Video  & Cls. & Other videos & \xmark & \xmark & 48,190 & 104,500 & \cmark \\
Celeb-DF~\cite{Yuezun-CVPR2020} & 2020 & Video  & Cls. & Other videos & \xmark & \xmark & 590 & 5,639 & \xmark \\
DeeperForensics~\cite{Jiang-CVPR2020} & 2020 & Video & Cls. & Hired actors & \xmark & \xmark & 1,000 & 1,000 & \cmark \\
WildDeepfake~\cite{Bojia-MM2020} & 2020 & Image & Cls. & N/A & \xmark & \xmark & 0 & 707 & \xmark \\
\midrule
FFIW~\cite{Zhou-CVPR2021} & 2021 & Video & Det. & Other videos &  \cmark & \xmark  & 12,000  & 10,000  & \xmark  \\
OpenForensics~\cite{ltnghia-ICCV2021} & 2021 & Image & Det. / Seg. & GAN & \cmark & \cmark & 45,473 & 70,325 & \cmark \\
\bottomrule
\end{tabular}
}
\end{table}

Table~\ref{table:deepfake_datasets} shows the basic information for the main deepfake datasets. They can be categorized into two groups: those that support only image-wise or video-wise deepfake classification and those that support face-wise classification and localization tasks.

The DF-TIMIT dataset~\cite{Korshunov-2018} has 640 fake videos crafted from the Vid-TIMIT dataset~\cite{Sanderson-ICB2009} using Faceswap-GAN~\cite{faceswap-gan}. The UADFV dataset~\cite{Yang-ICASSP2019} consists of 98 videos, half of which are fake, created using FaceApp~\cite{faceapp}. The FaceForensics++ dataset~\cite{Rossler-ICCV2019} contains 1000 pristine videos from YouTube and 4000 synthetic videos manipulated using deepfake methods~\cite{deepfakes, Thies-CVPR2016, faceswap, Thies-TG2019}. The Google DFD dataset~\cite{google_dfd-2019} includes 3068 fake videos. The Facebook DFDC dataset~\cite{Dolhansky-2020} contains 128K original and manipulated videos created using various deepfake and augmentation methods~\cite{Ivan-2020, Huang-ECCV2012, Egor-ICCV2019, Nirkin-ICCV2019, Karras-CVPR2019}. The Celeb-DF dataset~\cite{Yuezun-CVPR2020} comprises 590 YouTube celebrity videos and 5,639 fake videos. The DeeperForensics dataset~\cite{Jiang-CVPR2020} consists of 10,000 videos manipulated using a deepfake VAE and augmentations on 1000 original videos in the FaceForensics++ dataset. The WildDeepfake dataset~\cite{Bojia-MM2020} contains face sequences extracted from 707 deepfake videos collected from the Internet.

The FFIW dataset~\cite{Zhou-CVPR2021} includes 10K videos, in which each video frame has three faces on average. Faces were randomly forged using three deepfake methods~\cite{Ivan-2020, Nirkin-ICCV2019, faceswap}. This dataset contains face-wise annotations for both deepfake classification and detection. The OpenForensics dataset~\cite{ltnghia-ICCV2021} has 115K images with more than 334K faces. Different from the other datasets, the forged faces in the OpenForensics dataset were infinitely synthesized using GAN models~\cite{Shen-CVPR2020, Pidhorskyi-CVPR2020}. The OpenForensics dataset also contains rich annotations that support multiple tasks.

\subsubsection{DFDC Challenge 2020} 

\begin{table}[t!]
\caption{Top-ranked teams in DFDC Challenge 2020~\cite{Dolhansky-2020}.}
\label{table:dfdc}
\centering
\begin{tabular}{|c|l|c|}
\toprule
\textbf{Ranking} & \textbf{Team} & \textbf{Overall Log Loss} \\
\midrule
1 & Selim Seferbekov & 0.4279 \\
2 & WM & 0.4284 \\
3 & NTechLab & 0.4345 \\
4 & Eighteen Years Old & 0.4347 \\
5 & The Medics & 0.4371 \\
\bottomrule
\end{tabular}
\end{table}


The DFDC Challenge 2020~\cite{Dolhansky-2020}\footnote{\url{https://ai.facebook.com/blog/deepfake-detection-challenge-results-an-open-initiative-to-advance-ai/}} was launched by Facebook in partnership with industrial and academic leaders to support the development of innovative technologies for detecting deepfakes and manipulated media. Participants had to identify whether a video contained a manipulated face. The DFDC Challenge was performed on 100,000 videos, making it the largest deepfake database to date. The challenge had 2,265 submissions on the leaderboard\footnote{\url{https://www.kaggle.com/c/deepfake-detection-challenge}}. 

Table \ref{table:dfdc} lists the top-ranked teams. The top-ranked team\footnote{\url{https://github.com/selimsef/dfdc_deepfake_challenge}} used a multi-task cascaded convolutional neural network (MTCNN)~\cite{Zhang-SPL2016} for face detection and then the EfficientNet-B7 for deepfake classification. A noisy student model pre-trained on the ImageNet dataset was fine-tuned using heavy augmentations. The prediction results for video frames were combined through a heuristic fusion module.

The second-ranked team\footnote{\url{https://github.com/cuihaoleo/kaggle-dfdc}} used an ensemble of the Xception classifier and two weakly supervised models for data augmentation to compute frame-by-frame results and then take the average of all frames.

The third-ranked team\footnote{\url{https://github.com/NTech-Lab/deepfake-detection-challenge}} used an ensemble of EfficientNet-B7 models. One model ran on frame sequences using 3D convolution layers, and the others ran frame-by-frame, differing in the size of the face crops and augmentations. A noisy student model pre-trained on the ImageNet dataset was fine-tuned using mixup augmentation~\cite{Zhang-ICLR2018}.

\subsubsection{DeeperForensics Challenge 2020}

\begin{table}[t!]
\caption{Top-ranked teams in DeeperForensics Challenge 2020~\cite{Jiang-2021}.}
\label{table:deeper_forensics}
\centering
\begin{tabular}{|c|l|c|}
\toprule
\textbf{Ranking} & \textbf{Team} & \textbf{Binary Cross-Entropy Loss} \\
\midrule
1 & Forensics & 0.2674 \\
2 & RealFace & 0.3699 \\
3 & VISG & 0.4060 \\
4 & jiashangplus & 0.4064 \\
5 & Miao & 0.4132 \\
\bottomrule
\end{tabular}
\end{table}


The DeeperForensics Challenge 2020~\cite{Jiang-2021}\footnote{\url{https://competitions.codalab.org/competitions/25228}} was hosted by Nanyang Technological University in conjunction with ECCV 2020. The challenge was aimed at promoting face forgery detection methods through evaluation of their performances on 60,000 videos generated by face-swapping frameworks. Of the 115 teams registered for the competition, 25 teams made valid submissions and were on the leaderboard\footnote{\url{https://competitions.codalab.org/competitions/25228\#results}}.  

Table \ref{table:deeper_forensics} lists the top-ranked teams. The top-ranked team\footnote{\url{https://github.com/beibuwandeluori/DeeperForensicsChallengeSolution}} used an MTCNN~\cite{Zhang-SPL2016} for face detection and an ensemble of the EfficientNet-B0, EfficientNet-B1, and EfficientNet-B2 models for face-wise classification. These deep-learning models were fine-tuned using many data augmentations.

The second-ranked team used RetinaFace~\cite{Deng-CVPR2020} to detect faces and then averaged the predicted results by using a video-based model and an image-based model. The image-based model took the median of the frame-wise prediction probabilities while the video-based model used an attention mechanism to fuse the temporal information between frames. Noisy student EfficientNet-B5 models pre-trained on the ImageNet dataset were fine-tuned using various data augmentations.

The third-ranked team combined all facial images detected using an MTCNN~\cite{Zhang-SPL2016} into a sequence and then predicted the results using a 3D CNN. Different networks (\ie, I3D~\cite{Carreira-CVPR2017}, 3D ResNet~\cite{Hara-ICCVW2017}, and R(2+1)D~\cite{Tran-CVPR2018}) trained on diverse data augmentations were used.

\subsubsection{OpenForensics Benchmark}

\begin{table}[t!]
\caption{OpenForensics benchmark for multi-face forgery detection and segmentation~\cite{ltnghia-ICCV2021}. Seen/Unseen stand for seen/unseen manipulated image generation methods.}
\label{table:benchmark_multiface}
\centering
\begin{tabular}{|l|c|c|c|c|c|}
\toprule
\textbf{} & \textbf{} & \multicolumn{2}{c|}{\textbf{Detection}} & \multicolumn{2}{c|}{\textbf{Segmentation}} \\
\cmidrule{3-6}
\textbf{Method} & \textbf{Year} & \textbf{Seen Images} & \textbf{Unseen Images} & \textbf{Seen Images} & \textbf{Unseen Images} \\
\midrule
Mask R-CNN~\cite{Kaiming-ICCV2017} & 2017  & 79.2 & 42.1 & 83.6 & 43.7 \\
MS R-CNN~\cite{Huang-CVPR2019} & 2019  & 79.0 & 42.2 & 85.1 & 43.3 \\
RetinaMask~\cite{Fu-2019} & 2019  & 80.0 & 48.5 & 82.8 & 48.0 \\
YOLACT~\cite{Bolya-ICCV2019} & 2019  & 68.1 & 49.4 & 72.5 & 51.8 \\
YOLACT++~\cite{Bolya-PAMI2020} & 2020  & 72.9 & 53.7 & 77.3 & 54.7 \\
CenterMask~\cite{Lee-CVPR2020} & 2020  & 85.5 & 0.03 & 87.2 & 0.02 \\
BlendMask~\cite{Chen-CVPR2020} & 2020  & 87.0 & 53.9 & 89.2 & 54.0 \\
PolarMask~\cite{Xie-CVPR2020} & 2020  & 85.0 & 51.7 & 85.0 & 52.7 \\
MEInst~\cite{Zhang-CVPR2020} & 2020  & 82.8 & 46.1 & 82.2 & 46.0 \\
CondInst~\cite{Tian-ECCV2020} & 2020  & 84.0 & 52.7 & 87.7  & 54.1 \\
\bottomrule
\end{tabular}
\end{table}

Different from existing deepfake datasets, which contain only a face in an image, the OpenForensics dataset~\cite{ltnghia-ICCV2021} is aimed at identifying forged faces among many real faces in an image. This aim is achieved by providing face-wise ground truths for all classification, detection, and segmentation tasks. In addition, the dataset provides two kinds of images, ones created using seen deepfake generation methods and ones created using unseen methods.

Table \ref{table:benchmark_multiface} shows the benchmark evaluation results for multi-face forgery detection and segmentation tasks on the OpenForensics dataset~\cite{ltnghia-ICCV2021}. For standard evaluation, BlendMask~\cite{Chen-CVPR2020}, a modern single-stage method with an attention mechanism, achieved the best performance on both the detection and segmentation tasks. The other modern single-stage methods (\ie, CenterMask~\cite{Lee-CVPR2020}, PolarMask~\cite{Xie-CVPR2020}, and CondInst~\cite{Tian-ECCV2020}) also had high performance.

As also shown in Table \ref{table:benchmark_multiface}, YOLACT++~\cite{Bolya-PAMI2020} and BlendMask~\cite{Chen-CVPR2020} were the most robust against unseen images. There was a substantial drop in performance for all methods, especially CenterMask~\cite{Lee-CVPR2020}, for which the accuracy dropped to nearly zero. Even the leading detection methods remain limited and cannot yet effectively address unseen images, \ie, those beyond the training set's distribution. 

To complete the evaluation of all deepfake detection tasks on the OpenForensics dataset, we conducted two additional experiments on recognizing the authenticity of cropped faces on both the attack and defense sides, which were not described in the original report~\cite{ltnghia-ICCV2021}.

In the attack experiment, we evaluated the performance of two widely used and high-performance deepfake classifiers (\ie, XceptionNet~\cite{Rossler-ICCV2019} and EfficientNet-B4~\cite{Dolhansky-2020}) against three commonly used gradient-based adversarial attacks~\cite{Rauber-2017} (the Fast Gradient Sign Method (FGSM)~\cite{Goodfellow-ICLR2014}, the Basic Iterative Method (BIM)~\cite{Kurakin-ICLRW2016}, and the Projected Gradient Descent (PGD) method~\cite{Madry-ICLR2018}). FGSM is a simple yet effective method for generating adversarial images. BIM is an improved version of FGSM—FGSM is repeated multiple times with small changes in the parameters to make the attack more effective. This improved method is called ``iterative-FGSM." PGD~\cite{Madry-ICLR2018} is also an iterative version of FGSM—random starts are used to generate adversarial disturbances that are less perceptible. 

We trained XceptionNet~\cite{Rossler-ICCV2019} and EfficientNet-B4~\cite{Dolhansky-2020} on the training set and evaluated them on the test-development set of the OpenForensics dataset using their public training parameters. Table \ref{table:op_benchmark_attack} shows the results for the attack methods. The performance of XceptionNet dropped more than $20\%$ when it was attacked by FGSM and by more about $90\%$ when it was attacked by BIM and PGD. Although EfficientNet achieved about the same accuracy as XceptionNet on original images, it was not as robust as XceptionNet when it was attacked by any of the adversarial methods: its accuracy was greatly degraded by $80\%$ to more than $90\%$. Hence, deepfake classifiers are quite vulnerable to adversarial attacks. 

\begin{figure}[t!]
    \centering
    \includegraphics[width=1\linewidth]{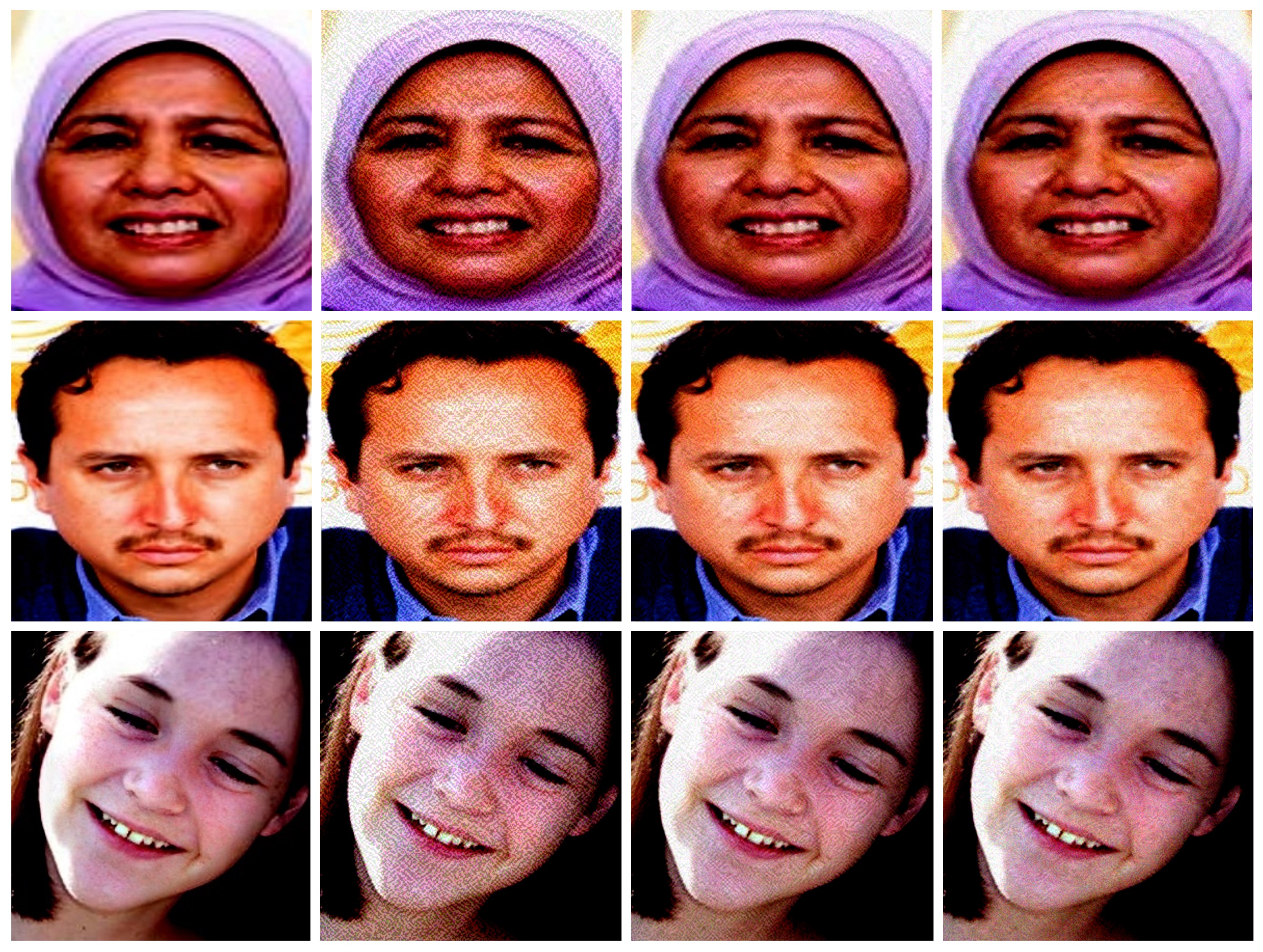}
    \caption{\highlight{Example images produced by compared adversarial attack methods. From left to right: original images followed by images produced by FGSM~\cite{Goodfellow-ICLR2014}, BIM~\cite{Kurakin-ICLRW2016}, and PGD~\cite{Madry-ICLR2018}. Best viewed in color with zoom.}}
    \label{fig:ae_visualization}
\end{figure}

\highlight{Figure~\ref{fig:ae_visualization} shows example images produced by the compared adversarial attack methods. FGSM was the worst in terms of both performance and realism. Both BIM and PGD had similar performance that was much better than that of FGSM.}

\begin{table}[t!]
\caption{Performance of deepfake classifiers against adversarial attacks evaluated on OpenForensics dataset~\cite{ltnghia-ICCV2021}.}
\label{table:op_benchmark_attack}
\centering
\begin{tabular}{|l|c|c|}
\toprule
\textbf{Adversarial Example} & \multicolumn{2}{c|}{\textbf{Victim Model}} \\
\cmidrule{2-3}
\textbf{Attack Method} & \textbf{XceptionNet~\cite{Rossler-ICCV2019}} & \textbf{EfficientNet-B4~\cite{Dolhansky-2020}} \\
\midrule
None & 99.22 & 99.58 \\
FGSM~\cite{Goodfellow-ICLR2014} & 76.98 & 19.76 \\
BIM~\cite{Kurakin-ICLRW2016} & 10.92 & 8.39 \\
PSD~\cite{Madry-ICLR2018} & 8.19 & 12.96 \\
\bottomrule
\end{tabular}
\end{table}

In the defense experiment, we evaluated the performance of defense solutions against seen and unseen scenarios. Particularly, we evaluated deepfake classifiers on the test-challenge set of the OpenForensics dataset, which was created by unknown generators. We trained XceptionNet~\cite{Rossler-ICCV2019}, EfficientNet-B4~\cite{Dolhansky-2020}, and CapsuleNet-R50~\cite{nhhuy-ICASSP2019} on standard images in the training set using their public training parameters. 

As shown in Table~\ref{table:op_benchmark_defense}, we evaluated them in two situations. First, under the assumption that the generators are seen, we re-used these methods to augment standard images in the training set for use in fine-tuning the deepfake classifiers. Using seen generators greatly improved the performance of the classifiers, to nearly $100\%$. Second, for unseen generators, we reconstructed their effects through unsupervised learning. In particular, we trained CycleGAN~\cite{Zhu-ICCV2017} on both standard images in the training set and unseen images in the test-challenge set, using public parameters. We then deployed CycleGAN on the training set to synthesize pseudo data for fine-tuning the deepfake classifiers to make them more robust. This increased the accuracy of all the classifiers by about $3\%$. This demonstrates that data augmentation can be used to train deepfake classifiers to make them more robust in both seen and unseen situations.

\begin{table}[t!]
\caption{Performance of defense solutions for deepfake classifiers against seen/unseen scenarios in OpenForensics dataset~\cite{ltnghia-ICCV2021}.}
\label{table:op_benchmark_defense}
\centering
\begin{tabular}{|l|c|c|c|}
\toprule
\textbf{Victim} & \multicolumn{3}{c|}{\textbf{Defense Solution}} \\
\cmidrule{2-4}
\textbf{Model} & \textbf{None} & \textbf{For Seen Generators} & \textbf{For Unseen Generators} \\
\midrule
XceptionNet~\cite{Rossler-ICCV2019} & 75.60 & 98.68 & 78.93 \\
EfficientNet-B4~\cite{Dolhansky-2020} & 83.89 & 93.67 & 85.99 \\
CapsuleNet-R50~\cite{nhhuy-ICASSP2019} & 73.07 & 95.99 & 76.30 \\
\bottomrule
\end{tabular}
\end{table}


\subsection{Future Directions}

\begin{figure}[t!]
    \centering
    \includegraphics[width=1\linewidth]{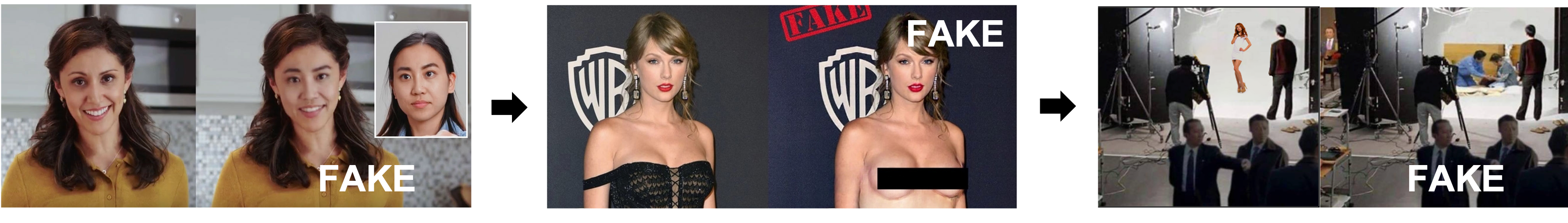}
    \caption{Predicted evolution of deepfake generation. From left to right, deepfake techniques will enable manipulation of human face, human body, and any visual content in the future. Note that DeepNude~\cite{deepnude} result (middle column) has been censored.}
    \label{fig:deepfake_evol}
\end{figure}

Deepfake generation methods are currently focused on the human face, so current deepfake detection methods are based on the assumption that the target objects to be manipulated are always faces. However, we cannot know which objects will be manipulated by attackers in the future because deepfake techniques can be applied to different visual objects. Indeed, DeepNude~\cite{deepnude} slightly modifies the Pix2PixHD GAN model~\cite{Wang-CVPR2018} to inpaint clothing areas with human skin, thus enabling anyone to be transformed non-consensually into a porn star. Thus, we believe that \textbf{deepfake techniques will be able to manipulate any visual content in the future}. We argue that this future manipulation will differ from conventional hand-crafted-based manipulation (\ie, adding, removing, and cloning objects). Figure \ref{fig:deepfake_evol} shows examples of different kinds of deepfakes. Hence, future generations of deepfake detection methods should have the \textbf{ability to identify} not only human faces but also \textbf{various manipulated visual contents}.

Existing deepfake detectors~\cite{nhhuy-BTAS2019, Li-CVPR2020, ltnghia-ICCV2021} \textbf{lack convincibility} because they show only possible forged regions in the images or video frames without any evidence, such as showing the original images before manipulation, namely fact verification. The problem is more challenging when the manipulation is adversarial and done using image filters in the digital world, so promoting efforts to detect deepfakes is very tricky for both people and machines.

\begin{figure}[t!]
    \centering
    \includegraphics[width=1\linewidth]{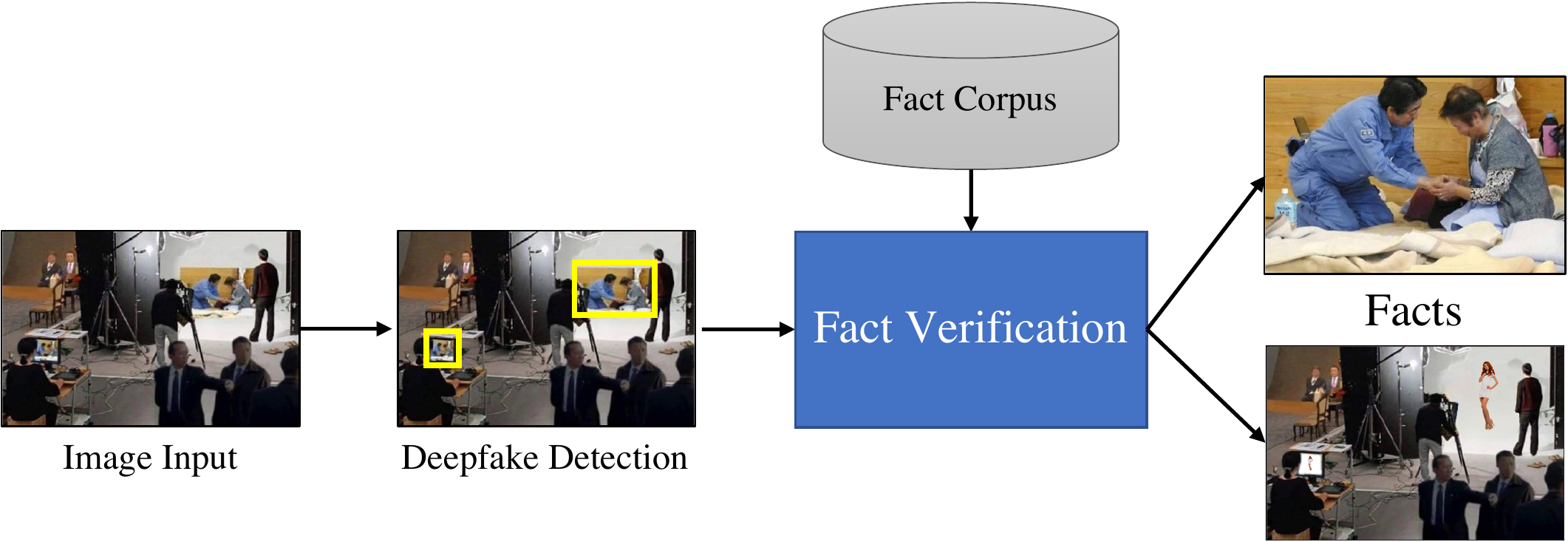}
    \caption{Suggested workflow for fact verification.}
    \label{fig:fact_verification}
\end{figure}

Verifying facts can help reduce false-positive results from deepfake detectors (authentic images or videos are classified as forged) and make them more reliable for real-world usage~\cite{Canasai-ACL2021} (e.g., journalism~\cite{Vo-2020}). Thus, image-based fact verification (\cf~Fig.~\ref{fig:fact_verification}) should be the focus in the next generation of deepfake detection methods, aiming towards trusted AI. However, to the best of our knowledge, current methods are unable to show the original data if they are manipulated.

\textbf{Fact verification} is a new and extremely challenging target for today's technologies. We argue that image-based face verification can be simulated via visual similarity search, \ie, copy detection~\cite{Douze-2021} and near-duplicate detection~\cite{Ke-MM2004}). In particular, assuming that a huge fact database can be crawled from the Internet, we can use a similarity search to find original-like images/videos in a fact database if they are manipulated. This research field has recently drawn attention from the research community through newly developed datasets~\cite{Heller-2018} and competitions~\cite{Douze-2021}. However, it still is a new problem without a clear state-of-the-art solution It is already particularly challenging to search through the huge volume of content currently being generated on social media; the volume of images and videos will likely reach into the millions or even billions.

\section{Summary} \label{section:conclusion}

This chapter overviewed deepfake generation and detection methods from the viewpoint of technical evolution in computer vision. In particular, it described deepfake generation methods in different categories and analyzed their limitations. In addition, it clarified the different tasks of deepfake detection, from conventional classification to modern end-to-end detection and segmentation. It further discussed the limitations of deepfake detection methods and suggested solutions for improving the robustness of deepfake detection. Finally, it suggested a future direction for deepfake detection.


\section*{Acknowledgments}

This research was partly supported by JSPS KAKENHI Grants (JP16H06302, JP18H04120, JP21H04907, JP20K23355, JP21K18023) and JST CREST Grants (JPMJCR20D3, JPMJCR18A6), Japan.


\bibliography{short_bibtex}
\bibliographystyle{IEEEtran}


\section*{Appendix}

This appendix lists links to the deepfake datasets and widely used methods described in this chapter. 


\begin{table}[h]
\centering
\begin{tabular}{l|l}
\multicolumn{1}{c|}{\textbf{Name}} & \multicolumn{1}{c}{\textbf{Link}} \\
 \midrule
OpenForensics~\cite{ltnghia-ICCV2021} & https://sites.google.com/view/ltnghia/research/openforensics \\
FaceForensics++~\cite{Rossler-ICCV2019} & https://github.com/ondyari/FaceForensics \\
FF++ Leaderboard & http://kaldir.vc.in.tum.de/faceforensics\_benchmark \\
DFDC~\cite{Dolhansky-2020} & https://ai.facebook.com/datasets/dfdc \\
DeeperForensics-1.0~\cite{Jiang-CVPR2020} & https://github.com/EndlessSora/DeeperForensics-1.0 \\
FFIW~\cite{Zhou-CVPR2021} & https://github.com/tfzhou/FFIW \\
XceptionNet~\cite{Rossler-ICCV2019} & https://github.com/ondyari/FaceForensics \\
EfficientNet-B4~\cite{Dolhansky-2020} & https://github.com/lukemelas/EfficientNet-PyTorch \\
CapsuleNet-R50~\cite{nhhuy-ICASSP2019} & https://github.com/nii-yamagishilab/Capsule-Forensics-v2 \\
CycleGAN~\cite{Zhu-ICCV2017} & https://github.com/junyanz/pytorch-CycleGAN-and-pix2pix \\
FoolBox~\cite{Rauber-2017} & https://github.com/bethgelab/foolbox
\end{tabular}
\end{table}

This chapter aims to introduce general knowledge about deepfake for beginners and/or students. A deeper understanding of the contents can be obtained by a review of the following materials for beginners. They should be helpful in obtaining basic knowledge about computer vision and biometrics, which is necessary for understanding deepfake.

\begin{table}[h]
\centering
\resizebox{1\linewidth}{!}{
\begin{tabular}{l|l}
\multicolumn{1}{c|}{\textbf{Name}} & \multicolumn{1}{c}{\textbf{Information}} \\
 \midrule
CVPR 2018 Tutorial on GANs & https://sites.google.com/view/cvpr2018tutorialongans \\
DCGAN Tutorial - Pytorch & https://pytorch.org/tutorials/beginner/dcgan\_faces\_tutorial.html \\
DCGAN Tutorial - Tensorflow & https://www.tensorflow.org/hub/tutorials/tf\_hub\_generative\_image\_module \\
DeepFaceLab Tutorial & https://medium.com/geekculture/creating-deepfake-miracles-with-\\&deepfacelab-tutorial-saehd-model-aa2aa12c08f3 \\
CVPR 2020 	
Workshop on & https://sites.google.com/view/wmediaforensics2020 \\ Media Forensics \\
Handbook of Biometrics & Jain, Anil K., Flynn, Patrick, Ross, Arun A., "Handbook of Biometrics", \\ & Springer, 2008. \\
\end{tabular}
}
\end{table}

\end{document}